\documentclass[10pt,twocolumn,letterpaper]{article}

\usepackage[cvpr]{cvpr}      

\usepackage{graphicx}
\usepackage{amsmath,mathtools}
\usepackage{amssymb}
\usepackage{booktabs}

\usepackage[pagebackref,breaklinks,colorlinks]{hyperref}

\usepackage[capitalize]{cleveref}
\crefname{section}{Sec.}{Secs.}
\Crefname{section}{Section}{Sections}
\Crefname{table}{Table}{Tables}
\crefname{table}{Tab.}{Tabs.}

\def\cvprPaperID{*****} 
\def\confName{CVPR}
\def\confYear{2022}

\begin{document}

\title{Controlling Structured Output Representations from Attributes using Conditional Generative Models}

\author{Mohamed Debbagh\\
McGill University\\
Montreal, Quebec \\
{\tt\small mohamed.debbagh@mail.mcgill.ca}
}
\maketitle

\begin{abstract}
Structured output representation is a generative task explored in 
computer vision that often times requires the mapping of low dimensional 
features to high dimensional structured outputs. Losses in complex spatial 
information in deterministic approaches such as Convolutional Neural Networks 
(CNN) lead to uncertainties and ambiguous structures within a single output 
representation. A probabilistic approach through deep Conditional Generative 
Models (CGM) is presented by Sohn et al. in which a  particular model known 
as the Conditional Variational Auto-encoder (CVAE) is introduced and explored. 
While the original paper focuses on the task of image segmentation, 
this paper adopts the CVAE framework for the task of controlled output 
representation through attributes. This approach allows us to learn a 
disentangled multimodal prior distribution, resulting in more controlled 
and robust approach to sample generation. In this work we recreate the 
CVAE architecture and train it on images conditioned on various attributes 
obtained from two image datasets; the Large-scale CelebFaces 
Attributes (CelebA) dataset and the Caltech-UCSD Birds (CUB-200-2011) dataset. We attempt 
to generate new faces with distinct attributes such as hair color and glasses, 
as well as different bird species samples with various attributes. We further
introduce strategies for improving generalized sample generation by applying 
a weighted term to the variational lower bound.

\end{abstract}

\section{Introduction}
\label{sec:intro}

Generative models have commonly been applied to computer vision tasks such 
as segmentation and image restoration problems. These methods often employ 
supervised approaches for learning feature representations on a set of images 
with the principal aim of performing an inference task over regions within an 
image. However, utilizing these learned representations to generate structured 
outputs can be challenging. Deterministic techniques such as 
Convolutional Neural Networks (CNN) perform fairly well, yet their generated 
outputs lead to ambiguous structures. The task of structured output 
representation often requires the mapping of the lower dimensional latent 
feature space to the high dimensional output space, in which spatial 
uncertainties come about as important features are lost during the encoding 
step of training. Moreover, deterministic models learn to map inputs to a single 
output representation which average out the variations in features,
leading to ambiguous structures in the generated output. 

A probabilistic approach can be taken toward describing these representations 
as a distribution of likely outcomes. The generative process of this approach  
captures variations in feature representations within a data distribution 
which materializes as a diverse, but likely, set of structured outputs. 
Thus, multiple outputs can be sampled from the distribution and further evaluated 
to select the desired outcome. One such probabilistic approach for structured output 
representation is the Variational Auto-encoder (VAE), which employs an auto-encoder 
framework to learn the parameters of a recognition model that is used to modulate 
a prior distribution on a latent variable \cite{kingma2022autoencoding}. 
VAEs allow us to generate structured output from a latent code by sampling 
from the prior distribution. However, this approach does not account 
for multiple modalities of the distribution, which can result in 
unconstrained generation of samples with little control of the variation of 
the outputs.

One approach that considers the modality of the data through a Conditional 
Generative Model (CGM) for the task of structured output representation is 
introduced by Sohn et al. \cite{NIPS2015_8d55a249}. This approach 
proposes the modulation of a conditional prior distribution of the latent 
space by learning a recognition model that is conditioned on an observation 
variable. At the time of sampling the (conditional) prior distribution, 
the outputs are constrained within the captured mode of the distribution.
The particular CGM proposed by the authors is known as the Conditional 
Variational Auto-encoder (CVAE) that builds upon the variational framework of 
the VAE to learn feature representations and conditions the generative model 
on input observation (e.g. labels, attributes, sensor data, partial images, etc.). 
This CGM approach allows us to sample from a multimodal distribution 
resulting in a more controlled and robust approach to sample generation. 
Sohn et al. implement the CVAE approach for the task of image segmentation and 
image restoration, in order to predict structured outputs generated by 
conditioning the models on the original images, with additional noise injected 
to improve training. In this work we explore the CVAE framework in producing disentangled 
structured output representations for the task of images synthesis from specified 
attributes. Rather than conditioning the CGM on the input image, our model will be 
conditioned on one-hot-encoded labels that describe the attributes of the subject of the original image. 
Performance of this model is tested on two image datasets; 
The first is the Large-scale CelebFaces Attributes (CelebA) 
dataset which describes faces with 40 attributes and the second is the Caltech-UCSD birds 
dataset which describes 200 bird species with various attributes. 

The sections of this paper are as follows. We review related works and key concepts, 
mainly from Sohn et al. \cite{NIPS2015_8d55a249}, in Section \ref{sec:related work} and \ref{sec:CVAE}.
This is followed by our implementation of the CVAE employed for 
image synthesis from attributes in Section \ref{sec:method}. 
Experiments to improve image synthesis are explored in Section \ref{sec:experiments}. 
This is  followed by the results and discussion in which we focus on 
mostly qualitative analysis of our CVAE implementation as 
comparisons to other approaches are beyond the scope of this research in 
Section \ref{sec:Results}. Finally, Section \ref{sec:Conclusion} concludes the paper.


\section{Related Work}
\label{sec:related work}

\subsection{Conditional Generative Models}
Deep Generative Models (DGM) are a class of directed graphical model-based approaches for structured 
output representation that aim to learn the underlying distribution 
of a data-set that can be sampled to generate new structured data representation. 
Prevalent generative models explored in the field of computer vision include 
Variational Auto-encoders (VAE) \cite{kingma2022autoencoding}, Generative 
Adversarial Networks (GAN) \cite{goodfellow2014generative}, 
Normalizing Flows \cite{rezende2015variational}, and 
Denoising Score Matching/ Diffusion Probabilistic Models \cite{song2021score}.

VAEs combine an auto-encoder framework with variational inference to learn a 
probabilistic mapping between the data and the latent space that is implicitly 
modelled with a known family of distribution (i.e. Gaussian). A latent code 
is then sampled and passed through a generator Neural Network (NN) to produce a 
structured output. GANs, on the other hand, consist of two NNs, a generator and 
a discriminator, that are trained simultaneously in an adversarial setting. 
The generator network is responsible for generating synthetic samples that 
closely resemble the real data. The generator takes random noise as input and 
transforms it into synthetic samples through a series of learned transformations. 
The discriminator network is a binary classifier that learns 
to distinguish between real samples from the dataset and fake samples generated 
by the generator. 
Normalizing Flows are generative models that learn an invertible 
transformation between the data distribution and a simple base distribution, 
such as a Gaussian or uniform distribution. This allows for exact computation 
of the likelihood and efficient generation of samples by inverting the 
transformation \cite{dinh2014nice, kingma2018glow}. 
Denoising Score Matching and Diffusion Probabilistic Models on the other hand learn to 
generate samples by simulating a stochastic process that gradually adds noise 
to the data until it reaches a simple base distribution. The generation 
process is then reversed by simulating a denoising process that removes the 
noise according to a learned score function \cite{ho2020denosing}.

Conditional Generative Models (CGMs) extend DGMs by conditioning the sample outputs on an additional 
input variable, such as observation data. This conditioning allows for more 
control over the structured outputs and enables the generation of samples within 
a specific modality of the output representation distribution. For each of the 
prevalent DGM mentioned, there are many works that incorporate a conditioned version
that constrain structured output to known states. The Conditional Variational Auto-encoder (CVAE) 
\cite{NIPS2015_8d55a249} extend the VAE framework by conditioning both 
the recognition and prior distribution models on the additional input variables. Similarly, 
Conditional GANs \cite{mirza2014conditional} extend the GAN framework by conditioning both 
the generator and discriminator networks on the additional input variables. 
Conditional Normalizing Flows have been proposed to enable the generation of 
samples conditioned on additional input variables. One such approach is the 
Conditional RealNVP \cite{dinh2016density}, which extends the RealNVP model by 
conditioning the affine coupling layers on the additional input variables. 
Conditional versions of Denoising Score Matching and Diffusion Probabilistic 
Models have also been explored, with the aim of learning to generate samples 
conditioned on additional input variables \cite{song2021score}.

\subsection{Disentangled Image Synthesis}
Disentangled image synthesis is a generative modeling task that involves 
generating new images that capture variations of the data representation 
in a controllable manner such that various attributes can be reproduced while 
maintaining other aspects of the image intact. Disentangled representations 
allows for better interpretability and control over the generated samples. 
This is challenging as variations can be ambiguous and difficult to identify.
There are several works that focus on disentangled image synthesis using 
different approaches to these generative frameworks.

In the context of VAEs, one such approach is to modulate the regularization 
term of the variational lower bound during the optimization process. 
\cite{higgins2017beta} introduces the $\beta$-VAE, which adds a hyperparameter, 
$\beta$, to the standard VAE objective function and acts as a weighted term 
that balances the trade-offs between the reconstruction loss and the latent space 
regularization. By increasing $\beta$, the $\beta$-VAE model puts more emphasis on the 
regularization term, which promotes the learning of independent 
factors of variation in the data. This leads to a more interpretable and 
controllable latent space, allowing for targeted manipulation of the generated samples. 
Similarly, in the context GANS, InfoGANs extends the standard GAN framework 
to achieve disintangled image synthesis by introducing an 
information-theoretic regularization term to the objective function to
maximize the mutual information between the input noise vector and the 
generated samples \cite{chen2016infogan}. This approach has been shown to be effective in unsupervised learning 
of interpretable latent factors and controlled generation of samples. 
Fader Networks \cite{lample2017fader} introduce an auto-encoder framework with 
an adversarial training scheme, where the encoder learns to remove specific 
attributes from the input images, and the decoder learns to add the attributes 
back. The disentangled attributes can be controlled through a latent code, which 
is concatenated with the content code obtained from the encoder.

\section{Preliminary: Conditional Variational Auto-Encoder}
\label{sec:CVAE}

In this section, we examine the underlying principles of the 
Conditional Variational Auto-Encoder (CVAE). The CVAE extends the fundamental 
mechanisms of the VAE, which employs a variational Bayesian approach. For each 
element of this method, we will first outline the VAE process and then discuss 
how the CVAE extends the baseline technique to incorporate conditionals.

\subsection{Auto-Encoding Variational Bayes Framework}
\begin{figure}[t]
    \centering
    \includegraphics[width=1\linewidth]{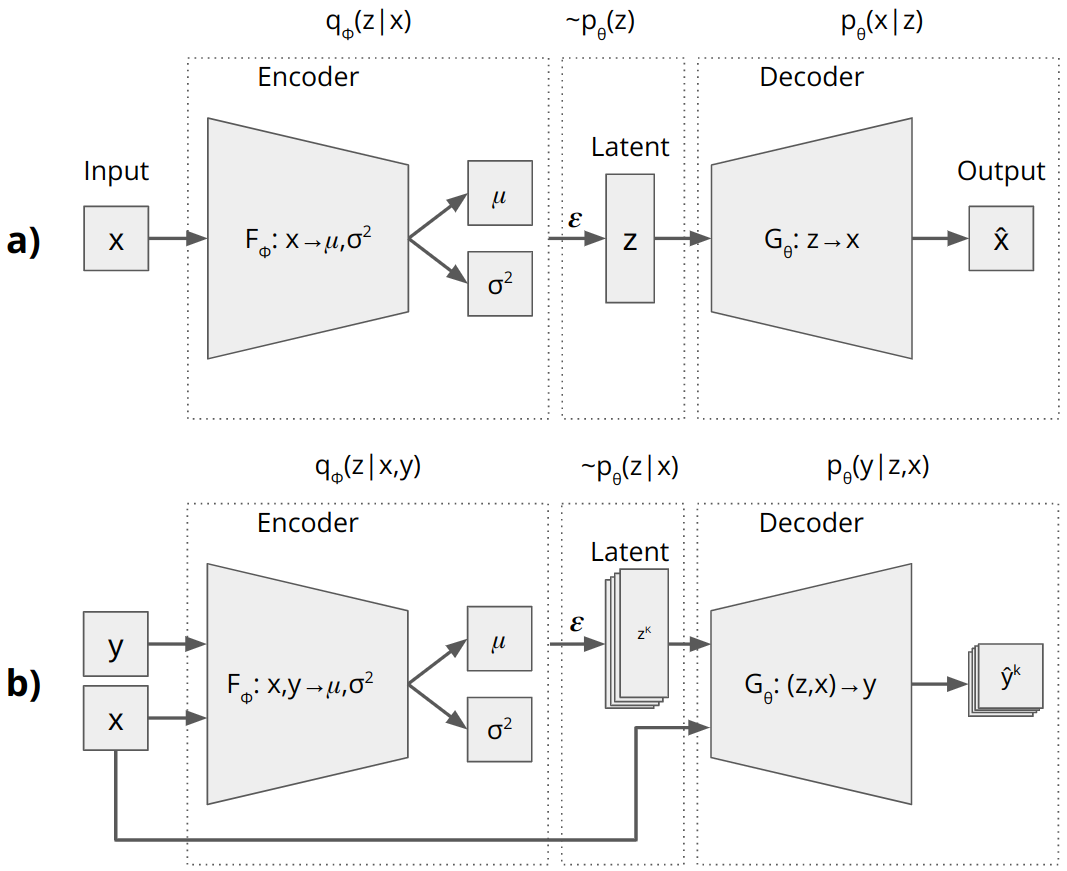}
  
     \caption{Overview framework of the a) VAE and b) CVAE. 
     This figure covers both deterministic and probabilistic components for each model.}
     \label{fig:framework}
  \end{figure}

Variational inference is used to approximate a posterior distribution of a 
directed graphical model whose latent variables and parameters are intractable. 
The Variational Auto-Encoder (VAE) combines this approach with an auto-encoder 
framework to learn the prior distribution of a latent space, $p_{\theta}(z)$, 
with parameters $\theta$. The idea is that the prior distribution can then be 
sampled to produce a latent code, $z$, which is passed as input to the decoder 
to produce a sample output, $\hat{x}$. VAEs consists of two NN for the 
probabilistic encoding and decoding process (see Figure \ref{fig:framework}a). 
As the true underlying distribution 
of the posterior is intractable and complex, a simple parametric surrogate 
distribution, $q_{\Phi}(z|x)$ (such as a Gaussian), with parameters $\Phi$, is assumed to 
approximate the distribution and is optimized for best fit. The encoder network implicitly models 
the surrogate distribution, by mapping the distribution parameters, $\Phi$, 
during the training process. The resulting model, $q_{\Phi}(z|x)$, 
is referred to as the recognition model. The optimization process of the 
recognition model revolves around minimizing the Kullback-Leibler (KL) divergence between the 
posterior and surrogate distributions, and subsequently maximizing the 
variational lower bound (see Section \ref{elbo}). Once the latent prior 
distribution is learned,  $z$ can be sampled via the reparameterization 
trick (see Section \ref{reparam}). The (probabilistic) decoder network performs 
a mapping of the latent code to a structured sample output for each sample, thus
producing a distribution of outputs, $p_{\theta}(x|z)$.

The CVAE expands upon the framework of the VAE, by combining 
variational inference with a conditional directed graphical model 
(see Figure \ref{fig:framework}b). In the case of 
CVAE, the objective is to learn a prior distribution of the latent space that is 
conditioned on an input variable $x$ such that $p_{\theta}(z|x)$. The prior can 
be sampled to generate a latent code, $z$ that is passed through the decoder with 
$x$ to produce a sampled output, $\hat{y}$. The recognition model in this case 
approximates the posterior distribution of the latent space given data, $y$, 
conditioned on $x$ such that $p_{\theta}(z|x,y)$. The conditioning of the distributions results in a prior that is 
modulated, by the input variable, creating a method to control modality of the 
output. The recognition model $q_{\Phi}(z|x,y)$ is optimized by minimizing the 
KL divergence between distributions $p_{\theta}(z|x,y)$ and $q_{\Phi}(z|x,y)$. 
$x$ and $y$ are passed as inputs to the encoder network during training to 
map the parameters, $\Phi$. Latent code, $z$ is sampled from the prior distribution 
via, the reparametarization trick, and passed as input along with $x$ to 
the decoder network, producing a distribution of outputs, $p_{\theta}(y|x,z)$.

\subsection{Variational Lower Bound} 
\label{elbo}
In Variational Inference, the main idea is to approximate the complex
intractable posterior distribution $p(z|x)$, which often arises in probabilistic 
graphical models and other Bayesian models containing latent variables, 
by using a simpler and tractable distribution, $q(z|x)$. This essentially, 
turns the task of inference into an optimization problem in which  
we minimize the Kullback-Leibler (KL) divergence between the true posterior 
and the approximate posterior as such:

\begin{equation}
    \begin{split}
        KL(q_{\phi}(z|x) || p_{\theta}(z|x)) =  
        \mathbb{E}_{q_{\phi}(z|x)}[\frac{\log q_{\phi}(z|x)}{\log  p_{\theta}(z|x)}  ].
    \end{split}
    \label{eqn:dkl}
\end{equation}

To overcome the issue of evaluating the intractable posterior in equation 
\ref{eqn:dkl}, we rewrite the KL divergence using Bayes theorem in terms 
of the marginal distribution $p_{\theta}(x)$ and joint distributions $p_{\theta}(x,z)$. 
We can further expand the equation in terms of the marginal log-likelihood (MLL):
\begin{equation}
    \begin{gathered}
        KL(q_{\phi}(z|x) || p_{\theta}(z|x))
        = \mathbb{E}_{q_{\phi}(z|x)}[\frac{\log q_{\phi}(z|x )p_{\theta}(x)}{\log  p_{\theta}(x,z)}] \\ 
        = \mathbb{E}_{q_{\phi}(z|x)}[\frac{\log q_{\phi}(z|x )}{\log  p_{\theta}(x,z)}] + \mathbb{E}_{q_{\phi}(z|x)}[ \log p_{\theta}(x)] \\
        = -\mathbb{E}_{q_{\phi}(z|x)}[\frac{\log  p_{\theta}(x,z)}{\log q_{\phi}(z|x )}] + \log p_{\theta}(x)
    \end{gathered}
    \label{eqn: mll}
\end{equation}

Now we notice that the terms in equation \ref{eqn: mll} follow particular 
properties. First, the KL divergence between $q(z|x)$ and $p(z|x)$ is always 
non-negative. We also notice that the second term is the MLL, and must be a 
negative value since $0<p(x)<1$. Thus, we can describe the first term on the 
right-hand side to be a lower bound of the marginal log-likelihood, and we can 
define it as such:

\begin{equation}
    \begin{multlined}
        \log p_{\theta}(x) \geq \mathcal{L}(\theta, \Phi)  = \\ 
        \mathbb{E}_{q_{\phi}(z|x)}[-\log q_{\phi}(z|x) + \log  p_{\theta}(x,z)]
    \end{multlined}
    \label{eqn: lb}
\end{equation}

As the variational lower bound, $\mathcal{L}(\theta, \Phi)$, increases and 
approaches the MLL, the KL divergence decreases, such that 
$KL(q_{\phi}(z|x) || p_{\theta}(z|x)) = 0$ when 
$\mathcal{L}(\theta, \Phi) = \log p_{\theta}(x)$. Thus, The optimization problem, can be evaluated as an objective function 
which aims to maximize the variational lower bound, which is also referred to as the
Evidence Lower BOund (ELBO).

The ELBO provided in equations \ref{eqn: lb} can be further expanded in terms of the 
expected log likelihood as such: 

\begin{equation}
    \begin{multlined}
        \mathcal{L}(\theta, \Phi)  = \\ 
        -KL(q_{\phi}(z|x) || p_{\theta}(z)) + \mathbb{E}_{q_{\phi}(z|x)}[p_{\theta}(x|z)]
    \end{multlined}
    \label{eqn: elbo}
\end{equation}

The second term on the right-hand side of equation \ref{eqn: elbo}, is the 
expected value of the log likelihood. The first term on the right-hand side is the KL 
divergence between the approximate posterior and the prior distributions. 

The ELBO, is a crucial aspect of optimizing the VAE and CVAE models. It serves as an 
objective function that needs to be maximized during the training process. 
The ELBO provides a balance between two competing terms: the reconstruction 
loss and the KL divergence. The reconstruction loss measures 
the difference between the original data and the data reconstructed by the 
model, ensuring that the model can generate accurate samples. On the other 
hand, the KL divergence term enforces a regularization constraint on the 
learned latent space, ensuring that the approximate posterior distribution 
remains close to the true prior distribution.

For the CVAE, the ELBO is slightly modified to account for the conditioning 
on the input observation variable $x$ and the reconstruction of data $y$:

\begin{equation}
    \begin{multlined}
        \mathcal{L}_{CVAE}(\theta, \Phi)  = \\ 
        -KL(q_{\phi}(z|x,y) || p_{\theta}(z|x)) + \mathbb{E}_{q_{\phi}(z|x,y)}[p_{\theta}(y|x,z)]
    \end{multlined}
    \label{eqn: cvae elbo}
\end{equation}

In equation \ref{eqn: cvae elbo}, the reconstruction loss term is adapted to include the conditioning on 
$x$, and the KL divergence term now measures the difference between the 
approximate posterior $q_{\Phi}(z|x,y)$ and the conditional prior 
$p_{\theta}(z|x)$. The optimization process of the CVAE revolves around 
maximizing this ELBO, resulting in an improved representation of the latent 
space and allowing for the generation of diverse and controlled outputs.

\subsection{Reparametrization Trick}
\label{reparam}
When modeling the recognition model implicitly by encoding the distribution 
parameters, the sampling step introduces non-differentiable operations that 
hinder the optimization process. In the context of VAEs, it is essential to 
have a differentiable method for generating samples from the approximate 
posterior distribution $q_{\Phi}(z|x)$. The reparameterization trick, 
proposed by Kingma et al. \cite{kingma2022autoencoding}, addresses the 
issue of non-differentiability in the latent space sampling step. This 
technique enables back propagation, an important algorithm for gradient-based 
optimization, to function effectively despite the stochasticity in the 
latent variable sampling process.

The reparameterization trick introduces an auxiliary noise variable, $\epsilon$, 
and reformulates the latent variable as a deterministic transformation. 
The noise variable is independent of the model parameters and input data 
with a standard normal distribution, i.e., $\epsilon \sim \mathcal{N}(0, I)$. 
This reformulation allows gradients to flow through the deterministic portion of the 
model, bypassing the random component. Consequently, this technique facilitates 
efficient gradient estimation, stabilizes the training process, and leads to 
improved convergence properties. 

The deterministic functional takes on the form, $z = g_{\Phi}(x,\epsilon), \epsilon \sim N(0,I)$, 
with parameters, $\Phi$, obtained from the encoder, typically taken as the parameters of 
a Gaussian distribution as such:

\begin{equation}
    \mu, \sigma^2 = f_{enc}(x; \Phi)
    \label{eqn:encoder_vae}
    \end{equation}

In the case of CVAEs, the reparameterization trick is applied similarly, but 
the encoder network is conditioned on both the input observation variable $x$ 
and the data to be reconstructed $y$:

\begin{equation}
\mu, \sigma^2 = f_{enc}(x, y; \Phi)
\label{eqn:encoder_cvae}
\end{equation}

The reparameterized latent variable for the CVAE is expressed as a 
deterministic transformation of the auxiliary variable, $z = g_{\Phi}(x,y, \epsilon)$. 
One such transformation, $g_{\Phi}(.)$, is given as:
\begin{equation}
z = \mu + \sigma \odot \epsilon, z \sim q_{\Phi}(z|x,y), \epsilon \sim N(0,I),
\label{eqn:reparam_cvae}
\end{equation}

The optimization process for the CVAE, like the VAE, centers around maximizing 
the ELBO, as depicted in equation \ref{eqn: cvae elbo}. By employing the 
reparameterization trick during the CVAE training process, the model is able to 
learn a more refined representation of the latent space, enabling the generation 
of diverse and controlled outputs based on the input observation variable $x$.

\section{Method}
\label{sec:method}
The Conditional Variational Auto-Encoder (CVAE) introduced in 
\cite{NIPS2015_8d55a249} was implemented for the task of image segmentation 
in a generative framework. As such, the authors condition their model on images 
as their input, $x$. This CVAE implementation is able to outperform 
equivalent deterministic methods, however, the task of image-to-image 
translation is often a transformation from one high dimensional space to 
another high dimensional space. This makes it easier for models to maintain 
spatial and structural information shared by the input and output image.
In this paper we evaluate a CVAE approach for the task of 
image synthesis from attributes, which performs a transformation from a low 
dimensional space to a high dimensional output space, specifically, 
text to images. This task is challenging as the generative network must construct 
structures that are not apparent from the low dimensional variables.
In this section, we present a CVAE implementation and performance enhancing 
methods for the task of attribute-to-image synthesis.

\subsection{Model Architecture}
\label{sec: model architecture}
\begin{figure*}[t]
    \centering
    \includegraphics[width=0.75\linewidth]{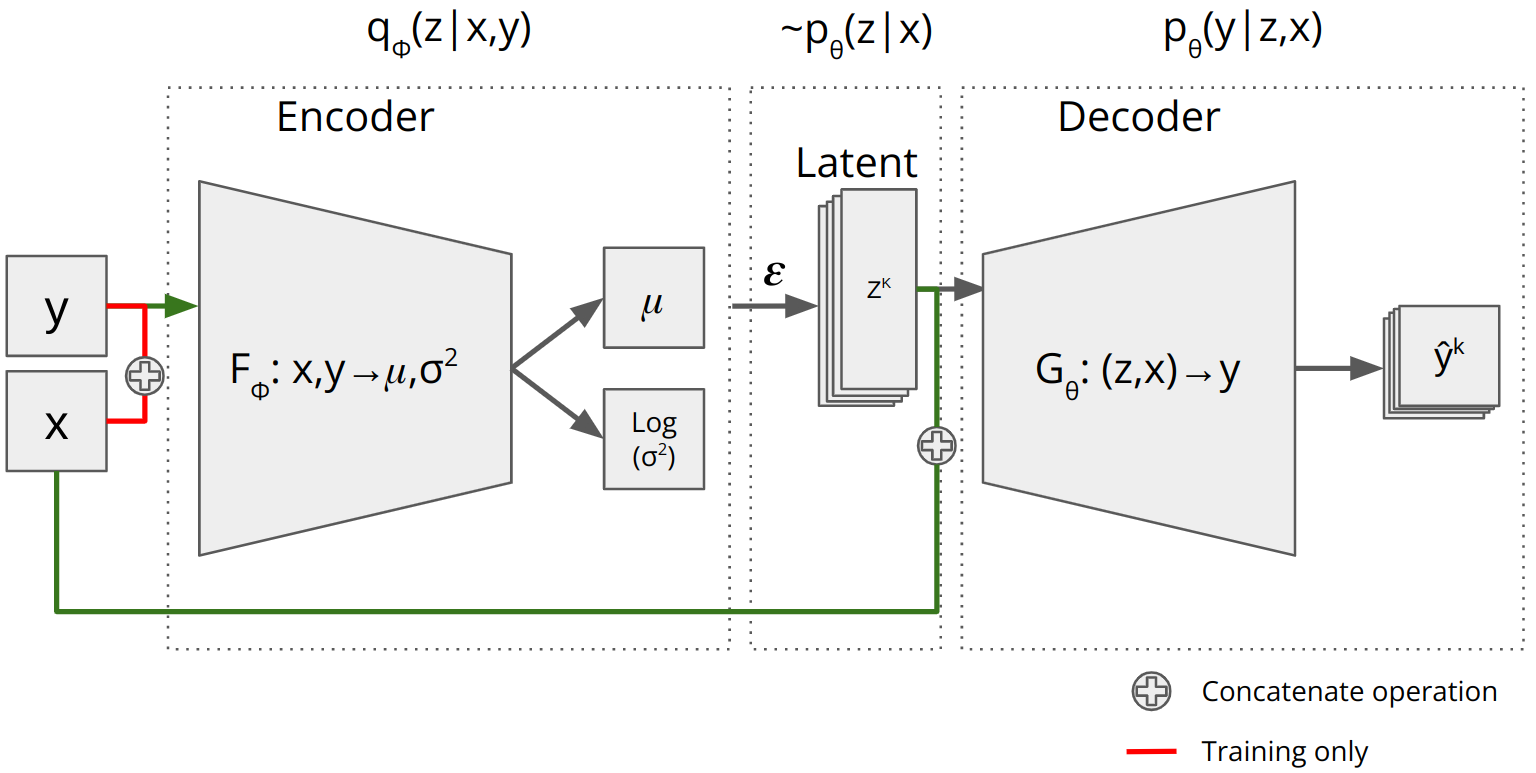}
  
     \caption{Overview of our CVAE architecture's training and sampling pipeline}
     \label{fig:architecture}
  \end{figure*}

Our model follows a standard CNN-CVAE architecture which consists of 
an encoder, a decoder, and a reparameterization step that samples 
the recognition model (see Figure \ref{fig:architecture}). The model is 
implemented in PyTorch \cite{paszke2019pytorch} and the main components 
of the CVAE model are described below.

\textbf{Encoder}  The encoder is a convolutional neural network (CNN) that 
takes both the training image $y$ and the conditioning variable $x$ as inputs. 
The conditioning variable $x$ is a one-hot-encoded tensor obtained from a 
list of string type attributes that correspond to training image, $y$. 
$x$ is reshaped and tiled to match the spatial dimensions of the training image, 
and then concatenated along the channel axis. The encoder processes the concatenated 
input through two convolutional layers, each followed by a ReLU activation 
function and max-pooling operation. The output is then flattened and passed 
through two fully connected Neural Network layers to obtain the mean $\mu$ and 
the log-variance $\log \sigma^2$ of the latent space. Encoding the log-variance 
provides a more stable, unconstrained optimization landscape and simplifies 
the loss function's gradients, making it easier for the model to learn and 
optimize.

\textbf{Reparameterization} The reparameterization step is responsible for 
generating samples of latent code, $z$ from the approximate posterior 
distribution $q_{\Phi}(z|x)$. 
It takes the mean $\mu$ and the log-variance $\log \sigma^2$ as inputs, and applies 
the reparameterization trick. The log-variance is first transformed into 
standard deviation, and an auxiliary noise variable $\epsilon$ is sampled from 
a standard normal distribution, $N(0,I)$. The latent variable $z$ is then computed as 
$z = \mu + \sigma \odot \epsilon$.

\textbf{Decoder} The decoder is a CNN that takes the latent variable $z$ and 
the conditioning variable $x$ as inputs. 
The conditioning variable $x$ is reshaped and concatenated with the latent 
variable $z$. The combined tensor is passed through a fully connected layer NN, 
and then reshaped to match the spatial dimensions required for the transposed 
convolutional layers. The decoder processes the input through two transposed 
convolutional layers, each followed by a ReLU activation function, except for 
the final layer, which uses a Sigmoid activation to generate the output image, $\hat{y}$.

\textbf{Weight Initialization} The weights of the model are initialized using 
the Kaiming normal initialization, which is well-suited for ReLU activation 
functions \cite{he2015delving}. This initialization 
method is based on the insight that, in deep networks, it is important to 
maintain the variance of the activations and gradients across layers during 
both forward and backward passes. This helps in preventing vanishing or 
exploding gradients and speeds up the training process. The biases are further initialized 
to zero. 

The network architecture is kept shallow with only two convolutional layers 
and one linear layer for both the encoder and decoder. The first reason for 
this is for computational efficiency. Keeping the network shallow allows the 
training process to be faster and requires less computational resources, 
which is particularly important when working with large image datasets. 
Furthermore, using a shallow network helps avoid the vanishing 
gradient problem. The second reason for keeping the network architecture 
shallow is to prevent overfitting. Since the number of trainable parameters 
in a network increases with the number of layers, a deep network can be more 
prone to overfitting on the training data. Since one of the problems associated, 
with image synthesis with VAE's is the overfitting to training images, and overall 
bad generalization, it is import to help our model generalize in the unknown latent space.

It is also important to note that using a shallow network may not be able to capture complex 
relationships between features in the data and may not perform as well as a deeper 
network architecture. Therefore, the decision to use a shallow or deep 
network is based on the specific task on the datasets (see Section \ref{sec:experiments}) and 
the available computational resources(see Section \ref{sec:hardware_and_software_setup}).

\subsection{Loss function}
The optimization process for the CVAE revolves around maximizing the 
variational lower bound with the formulation presented in equation 
\ref{eqn: cvae elbo}. In order to achieve this, the loss function used for training is a 
combination of the reconstruction loss and the KL-divergence loss. The 
reconstruction loss is calculated as the mean squared error (MSE) between the 
output image, $\hat{y}$, and the original training image, $y$. The KL-divergence loss is calculated 
using the mean $\mu$ and the log-variance $\log \sigma^2$ produced by the encoder. 
The total loss is the sum of these two components. A weighted term, $\beta$, is introduced as a tunable parameter for 
controlling which term to place more importance. $\beta < 0.5$ corresponds to 
more importance placed on the reconstruction loss to the overall loss function, 
whereas $\beta>0.5$ corresponds to more importance placed on the regularizing 
term.

\begin{equation}
    \mathcal{L}_{CVAE} = (1-\beta) \cdot \text{MSE}(\hat{y}, y) + \beta \cdot \text{KL}(\mu, \log \sigma^2)
\end{equation}
where $\text{MSE}(\hat{y}, y)$ represents the mean squared error between 
the output and target images, and $\text{KL}(\mu, \log \sigma^2)$ is the 
KL-divergence term given by:
\begin{equation}
    \text{KL}(\mu, \log \sigma^2) = -\frac{1}{2} \sum_{i=1}^{n} \left(1 + \log \sigma_{i}^2 - \mu_i^2 - \text{exp}(\log \sigma_{i}^2 ) \right)
\end{equation}
Where, $n$ is the number of elements in the latent vector.

The CVAE model is trained using the Adam optimizer \cite{kingma2014adam}, a gradient-based optimization, with 
back propagation enabled by the reparameterization trick. The model architecture 
and implementation are designed to provide an effective and efficient way of 
learning a rich and diverse representation of the data in the latent space, 
while also enabling the generation of controlled and high-quality outputs.

\section{Experiments}
\label{sec:experiments}

\subsection{CelebFaces Attributes Dataset}

The CelebFaces Attributes Dataset (CelebA) \cite{liu2015faceattributes} is a 
large-scale face attributes dataset containing more than 200,000 celebrity images, each 
annotated with 40 binary attribute labels, such as gender, age, and hair 
color. The dataset consists of diverse images of various celebrities, making 
it a popular choice for facial attribute recognition and generation tasks. The 
images in CelebA are provided in a resolution of 178x218 pixels. These images 
are collected from various sources from the internet and do not have 
consistent background and are thus considered "in-the-wild". In our study, 
we employ the CelebA dataset to train our CVAE model on the task of 
image synthesis from attributes, focusing on generating images based on the 
provided facial attributes. No data augmentation or preprocessing is performed 
on these images for this experiment.

\subsection{Caltech-UCSD Birds-200-2011}
The Caltech-UCSD Birds-200-2011 (CUB-200-2011) dataset \cite{WahCUB_200_2011} is a 
widely used dataset for fine-grained visual categorization. It consists of 
11,788 images of 200 different bird species. Each image is annotated with a 
bounding box, part locations, and 312 binary attribute labels, making it suitable for 
various computer vision tasks such as object detection, segmentation, and 
fine-grained classification. The dataset is split into a training set containing 
5,994 images and a test set containing 5,794 images. In our experiments, 
we utilize the bird images and their corresponding attribute labels to train 
and evaluate our CVAE model. In this experiment bird species are treated as 
additional attributes, and are converted to into one-hot-encoding and 
concatenated with the rest of the binary attributes. Images are taken from the internet 
and are considered to be "in-the-wild". Sizes of each image vary and are resized to 244x244px, 
before feeding through our models.

\subsection{Effect of $\beta$ on Regularization}
To investigate the impact of the $\beta$ parameter on the balance between 
reconstruction and regularization, we trained the CVAE model using different 
values of $\beta$ ranging from 0.25 to 0.9. The values were chosen to provide 
insights on the behavior of the model with varying levels of importance assigned 
to the KL-divergence vs reconstruction terms in the loss function. The $\beta$ 
values tested were 0.25, 0.5, 0.75, and 0.9. The CVAE model was trained separately for each $\beta$ value, keeping all other 
hyperparameters consistent across the experiments. The learning rate, batch 
size, and number of epochs were kept constant for a fair comparison. The CVAE 
loss function was used to compute the balance between reconstruction and 
regularization. After training the model with different $\beta$ values, the quality of the 
reconstructions, the latent space representations, and the generalization 
capability of the model were analyzed using the following evaluation metrics:

\textbf{Reconstruction Error} The reconstruction error for each $\beta$ value 
is computed by measuring the mean squared error (MSE) between the input images 
and the reconstructed images. This metric helps in understanding how well the 
CVAE model can reproduce the input data.

\textbf{Generalization and Disentanglement} the latent space 
is evaluated by analyzing the relationship between the latent variables and 
the semantic attributes of the input data. Generalization of the model 
is assessed by observing its performance on unseen data. This was achieved by 
evaluating the attributes present in images with varying attributes, as well as, 
assessing the quality of image generated. A more disentangled latent space 
would lead to better interpretability and generalization.

\subsection{Hardware and Software Setup}
\label{sec:hardware_and_software_setup}

The experiments in this study were conducted using a workstation equipped 
with an NVIDIA GeForce RTX 2080 Ti GPU, an AMD Ryzen 7 2700X Eight-Core 
Processor, and 64 GB of RAM. The operating system installed on the workstation 
was Ubuntu 22.04.2 LTS. For the implementation of the model and conducting 
experiments, we utilized several software libraries. PyTorch version 2.0, 
compiled with CUDA 11.7 for GPU acceleration, was employed for the deep 
learning framework. Additionally, NumPy was used for numerical computation and 
data manipulation, while Matplotlib was utilized for data visualization and 
generating plots.

\section{Results and Discussion}
\label{sec:Results}
\subsection{Synthesized Image Assessment}
\begin{figure*}
    \centering
    \includegraphics[width=1\linewidth]{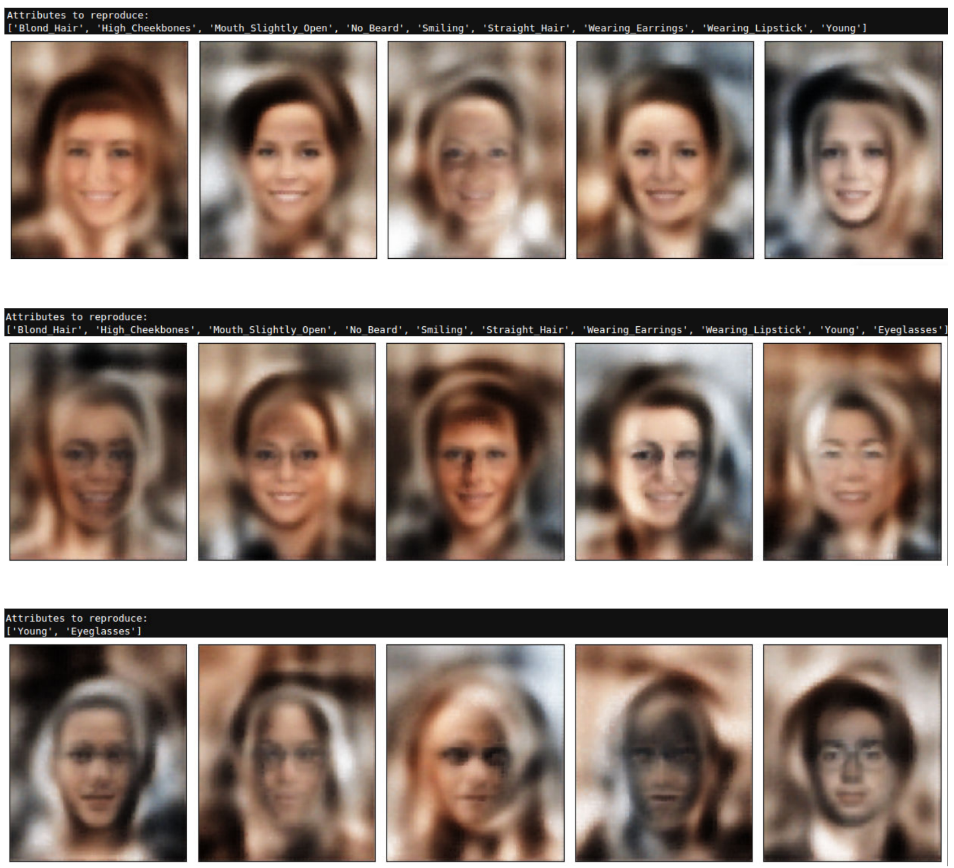}
  
     \caption{Samples generated to demonstrate robustness of the CVAE mdoel}
     \label{fig:playground}
  \end{figure*}

\textbf{CelebA} A qualitative assessment shows that our CVAE architecture is 
able to generate structured output that resemble human faces with the desired 
attributes. In general, facial structures are constructed from a set of string 
based attributes and a randomly sampled latent code, without any other 
information about spatial interactions. We notice that faces from the training 
data are typically found towards the center of the image and all faces tend 
to hold similarities in spatial structures. For example, a nose, two eyes, a 
mouth and general shape tend to have spatial consistency. More variability 
is introduced as we move further from the center, attributes such as 
hairstyles, outfit and background, change significantly and often are not 
captured in the list of attributes. This is reflected in the generated images, 
as facial structures are captured in higher detail in the center, and as 
we move further out, structures tend to be more ambiguous and blurry. To test 
the robustness of the model, random attributes were selected, added and removed 
to measure the extent of the samples generated, and indeed fairly good 
samples are generated for all cases (see Figure \ref{fig:playground}).
 
\begin{figure}[t]
    \centering
    \begin{subfigure}{1\linewidth}
        \includegraphics[width=1\linewidth]{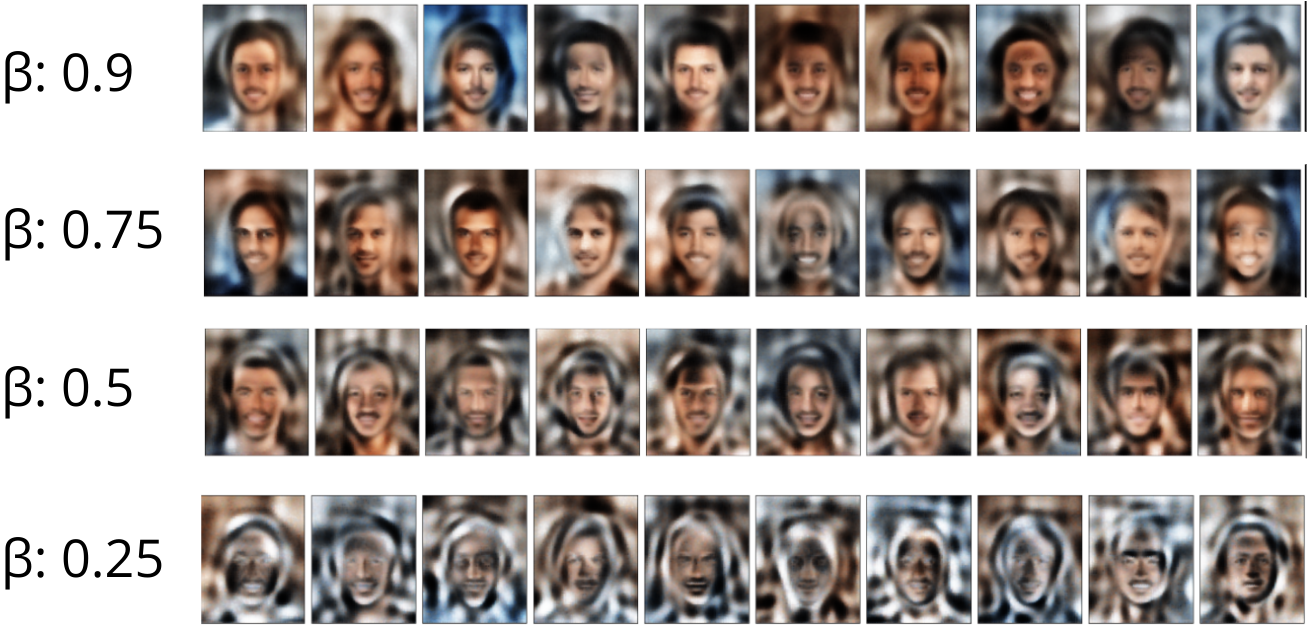}
        \caption{CelebA}
        \label{fig:celeba results}
    \end{subfigure}
    
    \begin{subfigure}{1\linewidth}
        \includegraphics[width=1\linewidth]{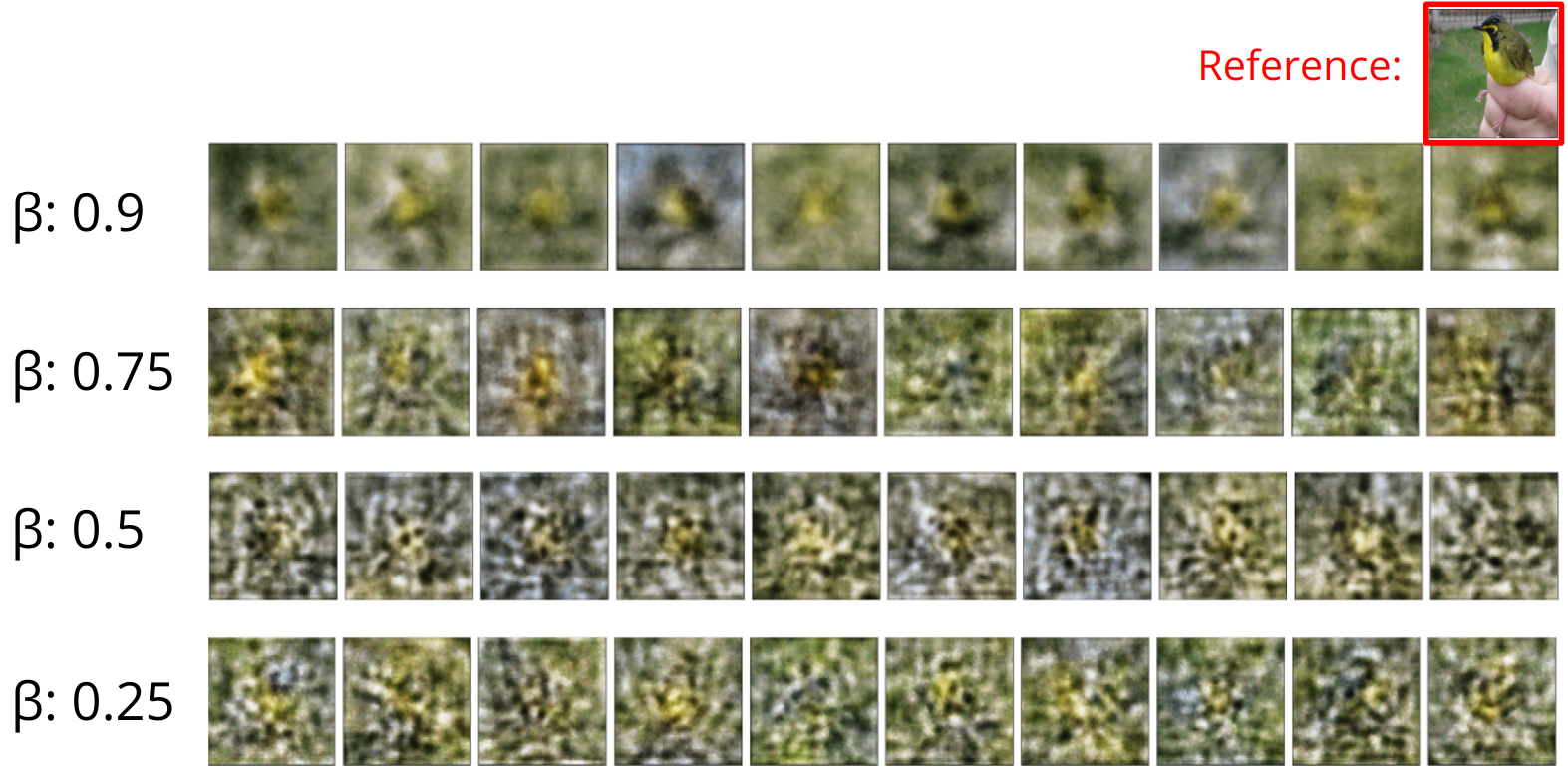}
        \caption{CUB-200-2011}
        \label{fig:birds results}

    \end{subfigure}
    \label{fig:beta_output}
    \caption{Samples generated from model trained on $\beta$ values ranging from 0.25 to 0.9}
\end{figure}

\textbf{CUB-200-2011} A qualitative assessment of the samples show that we 
were not able to generate highly detailed structured images of birds. However, 
we notice that the model tries to recreate generalities about the attributes that 
birds hold. If we examine in Figure \ref{fig:birds results} our model is able to 
generally, produce the color palette of the bird with desired attributes. If we 
further examine the first row, as we increase regularization, more structures 
seem to be apparent, albeit blurry. This is most likely due to the variability 
in structures within the images, as well as sparsity of training images with the held 
attributes.

\subsection{Reconstruction vs Regularization}
\begin{figure*}[hbt!]
    \centering
    \begin{subfigure}{1\linewidth}
        \includegraphics[width=.75\linewidth]{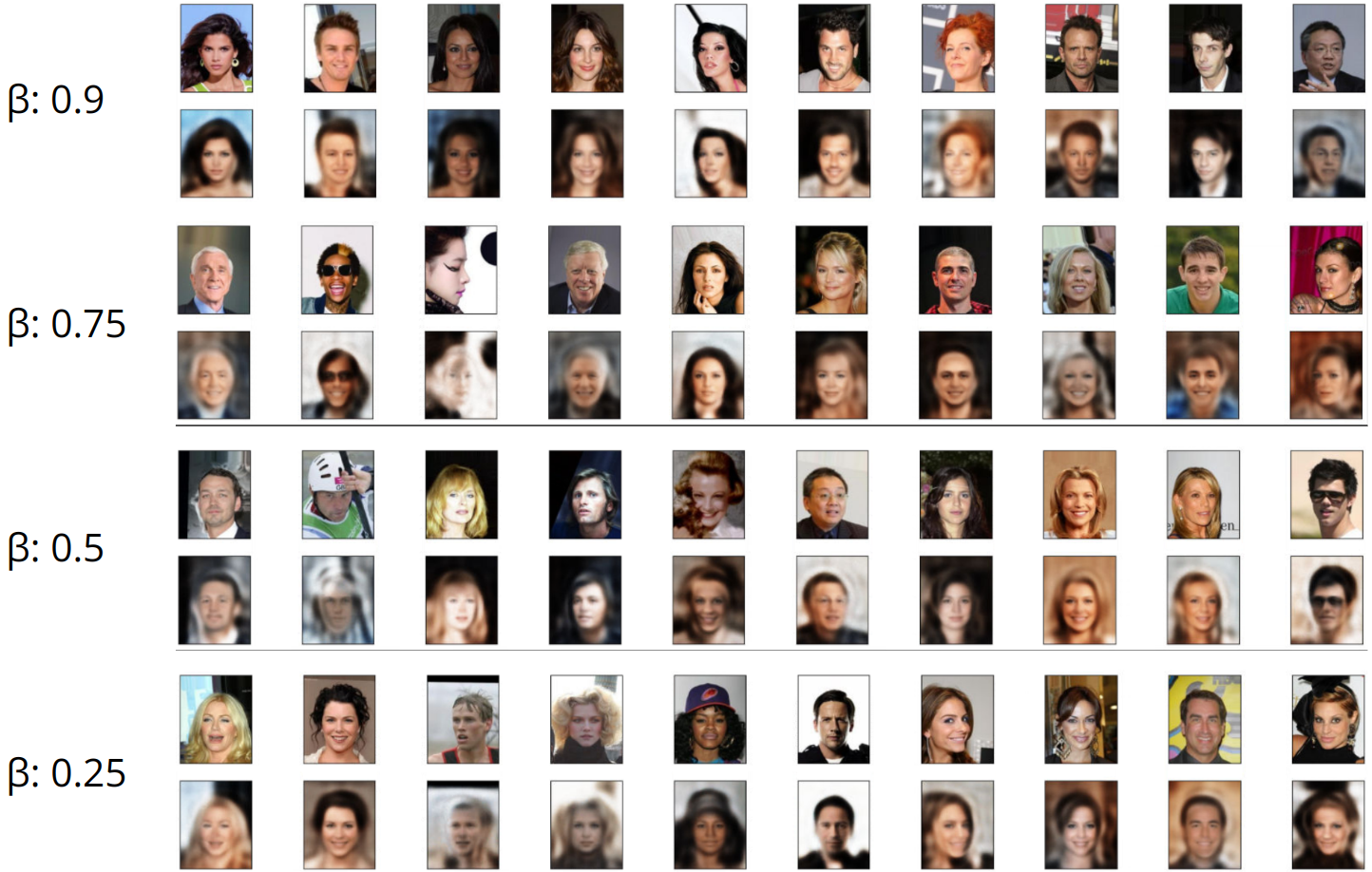}
        \caption{CelebA}
        \label{fig:celeba reconstruction}
    \end{subfigure}
    
    \begin{subfigure}{1\linewidth}
        \includegraphics[width=.55\linewidth]{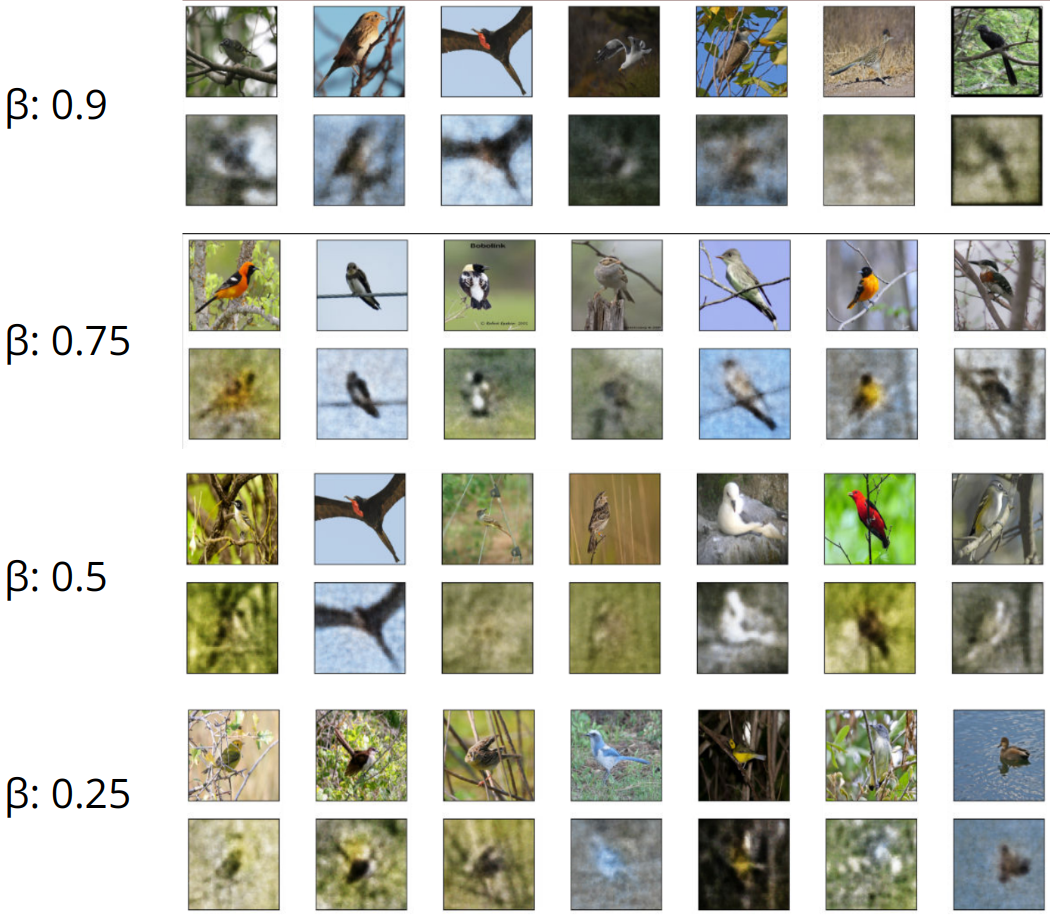}
        \caption{CUB-200-2011}
        \label{fig:birds reconstruction}

    \end{subfigure}
    \label{fig:beta_recon}
    \caption{Reconstructed images generated during training from model with $\beta$ values ranging from 0.25 to 0.9. For each $\beta$ the upper row is the Ground Truth image and the bottom is the reconstructed image}
\end{figure*}

The intuition behind modulating the regularization and reconstruction terms 
is to penalize overfitting of the recognition model to the structures of the 
training data, while maintaining the structural learning process. We notice in 
Figures \ref{fig:celeba results} and \ref{fig:birds results}
that low $\beta$ value, result in high frequency ambiguities (in celebA) and 
artifacts (in cub-200-2011) when generating samples. While we are able to achieve 
very accurate reconstructions of the training images in both datasets, 
the objective is to synthesize a variety of generalized faces that hold 
those attributes. Thus, querying the generative network trained on a low 
$\beta$ value results in uncertain and ambiguous structures in the output image. As we increase $\beta$, 
however, the regularization term is favored, resulting in much more natural generation 
of the faces, and smoother and structurally sound images of birds. In Table \ref{tab:beta} we observe 
that $\beta = 0.25$ achieves the lowest mean squared error, which corresponds to 
accurate reconstructions during training (see Figure \ref{fig:celeba reconstruction}). However, 
it is apparent that sampling from selected attributes and a random latent code, 
the network trained on  $\beta = 0.25$, performs the worst, with unrealistic synthesized images, 
as show in the last row of Figure \ref{fig:celeba results}. This is especially 
apparent in the CUB-200-2011 dataset, in Figure \ref{fig:birds reconstruction}, as the model 
does a very good job at reconstructing the training image, however it fails to generate 
any significant structures during sampling. Since the task of image synthesis 
is to generate variations of structured outputs from the same attributes, a 
higher $\beta$ value is desired, as it helps the model avoid overfitting and puts more 
significance in achieving a regularized solution that keeps our recognition 
model close the prior distributions of the latent space.

\begin{table}
    \centering
    \begin{tabular}{c |c c | c c}
      \toprule
      &\multicolumn{2}{c|}{CelebA}&\multicolumn{2}{c}{CUB-200-2011}\\
      $\beta$ & MSE & KL & MSE & KL \\
      \midrule
      0.9 & 82787 & 9515 & 20129 & 482\\ 
      0.75 & 81157 & 10033 & 13446 & 2315\\   
      0.5 & 73137 & 15845 & 13256 & 3677\\  
      0.25 & 70444 & 27107 & 12093 & 4957\\  
      
      \bottomrule
    \end{tabular}
    \caption{Mean Squared Error and KL losses for models trained on $\beta$ ranging from 0.25 to 0.9}
    \label{tab:beta}
\end{table}

\section{Conclusion}
\label{sec:Conclusion}
In this paper, we have demonstrated the effectiveness of 
the Conditional Variational Auto-Encoder as a 
powerful generative model for constructing high-dimensional 
structured outputs from low-dimensional attributes. This 
framework enables the generation of disentangled samples of 
images with numerous variations of the same attributes, 
highlighting its potential for diverse applications in image 
synthesis. Our study further reveals that a higher weight placed on the 
regularization term of the loss function of the CVAE during training is 
favored in image synthesis tasks. This is because the 
primary objective is not to learn accurate 
reconstructions but rather to capture the variations in the 
latent space effectively. However, challenges remain in 
capturing high-frequency variabilities in sparse image 
datasets. Despite these limitations, our CVAE model still 
demonstrates the ability to generate meaningful 
representations in a controlled manner. Image synthesis from text is 
often challenging, as spatial information is not directly 
available to the model. The CVAE framework 
proves its versatility and adaptability to such complex 
scenarios.

Finally, improvements to data pre-processing and data augmentation techniques
can significantly improve results. As a fully supervised method, CVAEs 
are completely dependent on the quality of the data it is trained on. 
As such various challenges arise, especially when dealing with uncertainties and implicit biases. 
Further research should be conducted in this area to improve utilization of deep Condition Generative Models, 
such as the CVAE.

{\small
\bibliographystyle{ieee_fullname}
\bibliography{main}
}
\clearpage

\end{document}